\documentclass[11pt]{article}

\usepackage{natbib}
\setcitestyle{numbers}
\usepackage{authblk}
\usepackage[margin=1in]{geometry}
\usepackage{bbm}
\usepackage[utf8]{inputenc} 
\usepackage[T1]{fontenc}    
\usepackage{hyperref}       
\usepackage{url}            
\usepackage{booktabs}       
\usepackage{amsfonts}       
\usepackage{nicefrac}       
\usepackage{microtype}      
\usepackage{amsmath}
\usepackage{amssymb}
\usepackage{mathtools}
\usepackage{amsthm}
\usepackage{xcolor}
\usepackage{arydshln}
\usepackage{subcaption}
\usepackage{caption}

\author{Jiachen Zhao}
\affil{University of Massachusetts Amherst\\
{\tt jiachenzhao@umass.edu}}

\date{}

\begin{document}
\title{In-Context Exemplars as Clues to Retrieving from Large Associative Memory}
\maketitle





\begin{abstract}
Recently, large language models (LLMs) have made remarkable progress in natural language processing. The most representative ability of LLMs is in-context learning (ICL), which enables LLMs to learn patterns from in-context exemplars without training.  The performance of ICL greatly depends on the exemplars used. However, how to choose exemplars remains unclear due to the lack of understanding of how in-context learning works. In this paper, we present a novel perspective on ICL by conceptualizing it as contextual retrieval from a model of associative memory. We establish a theoretical framework of ICL based on Hopfield Networks.  Based on our framework, we look into how in-context exemplars influence the performance of ICL and propose more efficient active exemplar selection.  Our study sheds new light on the mechanism of ICL by connecting it to memory retrieval, with potential implications for advancing the understanding of LLMs.
\end{abstract}


\section{Introduction}
In recent years, large language models (LLMs) have garnered significant attention due to their ability to revolutionize natural language processing (NLP) by demonstrating impressive language understanding and reasoning capabilities~\citep{bubeck2023sparks,brown2020language,singhal2022large,wei2022chain,shin-etal-2021-constrained}.  LLMs are first pretrained on extensive data using the language modeling technique where the model predicts the next token given a context.  Without finetuning on task-specific data, LLMs leverage in-context learning (ICL), also referred to as few-shot prompting, to make predictions. Through ICL, LLMs can find underlying patterns of the input query through given in-context exemplars, such as a set of input/output pairs, and use them to complete the response.

However, the effects of in-context exemplars on downstream performance via ICL and guidelines for formulating those exemplars (e.g., how to select exemplars and how many exemplars to use) remain unclear. Because the understanding of how ICL works is currently intuitive and lacks theoretical foundation. Past works to understand ICL mainly focus on empirical investigation~\citep{min-etal-2022-rethinking,garg2022can,liu-etal-2022-makes,DBLP:conf/emnlp/YooKKCJLLK22}.  Therefore, this work are motivated to build a theoretical framework to analyze ICL and then to provide implications to formulating in-context exemplars for stronger performance. More specifically, we adopt a novel and distinct perspective by conceptually reframing ICL as contextual \textit{retrieval} rather than a \textit{learning} problem~\citep{dai2022can,von2022transformers,dong2022survey}, as there is no actual weight update involved. We conceptualize LLM as a biologically plausible model of associative memory~\citep{hopfield1982neural}, also known as content-addressable memory.  In psychology, associative memory refers to the aptitude for linking and recollecting numerous sets of unconnected items, also known as memories. If furnished with a subset of items derived from a specific memory, an organism possessing associative memory can retrieve the remaining items associated with that particular memory. Such memory retrieval in human brains is also triggered by cue/ prompt. Contextual cues are essential to successful memory recall in brain and increase the accessibility of memory~\citep{tulving1973encoding}.

In the realm of machine learning, memory models~\citep{hopfield1982neural,kanerva1988sparse,sukhbaatar2015end,kaiser2016can,ramsauer2020hopfield, krotov2016dense} have been widely studied for a long time. Two fundamental models are Hopfield Network~\citep{hopfield1982neural} and its extension, sparse distributed memory~\citep{kanerva1988sparse}. Generally, the memory retrieval process in these models is executed by updating neuron configurations through a predetermined rule to minimize the network's energy. This retrieval process can also be viewed as pattern recognition~\citep{krotov2016dense}. An input is perceived as a query related to patterns stored in the model's memory, and the model generates a prediction by recalling its memory based on the query. 

Through the lens of memory models, we demonstrate that ICL with self-attention~\cite{vaswani2017attention} in LLMs can be interpreted as retrieving patterns from {associative memory} of Hopfield Networks with context.  Correspondingly, we establish a theoretical framework and analyze retrieval error, i.e., downstream performance via ICL.  Within our contextual retrieval framework, we look into the influence of in-context exemplars on performance of ICL both theoretically and empirically. We justify why randomly choosing exemplars can work especially given enough exemplars. Our analysis also suggests that different from supervised learning, ICL will not necessarily have better performance with more exemplars, which depends on the chosen exemplars. Moreover, we further propose efficient Active Exemplar Selection which achieves better performance with much fewer exemplars than random selection.  

\section{Brief Review of Hopfield Networks}\label{app:hopfield}
Hopfield Networks (HNs)~\cite{hopfield1982neural} were introduced to store and retrieve information. The standard HN~\cite{hopfield1982neural} consists of a neural network of $N$ neurons that can in total store $M$ binary patterns of dimension $D$.  Memory $\xi$ is denoted as an array of stored pattern vectors, i.e., $\xi=[m_{1},...,m_{M}]$, where $\xi\in\mathbb{R}^{M\times D}$. During the retrieval process, the configuration of neurons is fixed to the query pattern (e.g., incomplete $m_{i}$), and an update rule $f$ for $\sigma$ is defined to retrieve the similar or the same pattern to the query. Each update lowers the energy function $E$ of the network, which belongs to the Ising spin-glass model~\cite{scott1978infinite} in physics. The energy is expected to converge to an attractor state (local minimum) through repeated updates. Eventually, HNs will return the pattern from its memory that is the most similar to the input.  Additionally, HNs are similar to humans' memory system. The neuron's state corresponds to the firing rate or activity level of biological neurons. The weights of the network correspond to the strength of the synaptic connections between neurons in the brain. Similar to HNs, memories in brains are stored in a distributed manner across many regions and neurons. There are associative areas storing relations between features. Complex memories can then be recalled to generate predictions based on partial cues or associations~\cite{bonetti2021rapid,barron2020prediction} just like HNs.

\section{ICL as  Contextual Retrieval}
\label{sec:theo}
This section presents a formulation of ICL as pattern retrieval based on context from memories of modern Hopfield Networks (MHNs)\citep{ramsauer2020hopfield,krotov2016dense,millidge2022universal}.  Brief overview of Hopfield Networks is provided in Appendix~\ref{app:hopfield}.  In this section, we first give a formal setup of ICL. For a pre-trained language model whose parameters are denoted as $\theta$, given an input $\mathbf{x}$, the model will predict $\tilde{\mathbf{y}}$ for ground truth by conditioning on the query and a context sequence containing $K$ exemplars that are drawn from an accessible labeled dataset $\mathcal{D}_{(x,y)}$ (each exemplar $e_{i}=(x_{i},y_{i})$). Formally, we denote the sequence of all $K$ in-context exemplars $\mathbf{e}$, i.e., $\mathbf{e}=e_{1},...,e_{K}$. We can then have
\begin{equation}
    \tilde{\mathbf{y}}=\text{argmax}_{y}P(y|\mathbf{e},\mathbf{x},\theta).
\end{equation}

From the perspective of HNs, the input string [$\mathbf{e}$,$\mathbf{x}$] is a cue to the associative memory. The feed-forwarding process in the language model is to reconstruct the completion $\tilde{\mathbf{y}}$ of $\mathbf{x}$ that aligns with patterns of context $\mathbf{e}$ by recalling information stored into the model's memory during pretraining. For LLM, the pretraining is implemented as predicting masked/ next tokens of sentences, which is essentially teaching the model to reconstruct completion based on context like HNs.

To demonstrate the close relation between ICL and retrieving from HNs, we first extend the model definitions discussed by \citet{ramsauer2020hopfield, millidge2022universal} to construct a Hopfield Network with Context (HN-C). We then show that contextual retrieval from HN-C is equivalent to self-attention in LLMs.  To incorporate context in HNs, we consider stored patterns in memory as applying a linear transformation to raw vectors with a memory matrix, which is different from past frameworks \citep{ramsauer2020hopfield, krotov2016dense, millidge2022universal} that assume a static array of stored patterns. Thus, in our case, context patterns are dynamically defined depending on the input context. It is also important to note that retrieval does not necessarily mean extracting the exact stored patterns in memory without loss, but rather involves the induction of completion based on the input patterns that are typically not fully identical to the stored memories~\citep{krotov2016dense}. Actually, contextual retrieval setting is indeed how the human brain retrieves episodic memory~\citep{ranganath2002prefrontal}.

Formally, we denote some underlying query vector of input strings by $\sigma\in\mathbb{R}^{d_{m}}$. We define there are $M$ context vectors $\lambda_{i}\in\mathbb{R}^{d_{m}}$ which are represented by a matrix $\Lambda \in \mathbb{R}^{ d_{m}\times M}$.  We define memory matrix $\xi_{Q} \in \mathbb{R}^{ d_{m}\times d_{q}}$ and $\xi_{K} \in \mathbb{R}^{ d_{m}\times d_{q}}$ respectively for query vector $\sigma$ and context vector $\lambda$. We further define $\mathrm{Z}:=\xi_{K}^{T}\Lambda$ and each column vector $z$ is \textbf{context pattern}.   Accordingly, we have $u:=\sigma\xi_{Q}$ as \textbf{query pattern}.  We then define the update rule for $u$ of the model based on the Universal Hopfield Network~\cite{millidge2022universal} as follows: 
\begin{align}
       u^{\text{new}}&=sep (\gamma\, sim(u,\,\mathrm{Z}))\Lambda^{T}\xi_{K}\label{eq:1},
\end{align}
where $\gamma$ is a scalar value, $sim$ is a similarity function and $sep$ is a separation function. We set $sim$ as dot production and $sep$ as $softmax$ function. Then the update rule can be further specified as Eq.~\ref{eq:2}.  
\begin{align}
     u^{\text{new}}&=softmax(\gamma\,u\mathrm{Z})\Lambda^{T}\xi_{K}\\
     &=softmax(\gamma\,\sigma\xi_{Q}\xi_{K}^{T}\Lambda)\Lambda^{T}\xi_{K}\label{eq:2}
\end{align}
This formulation can be converted to self-attention by applying a linear transformation to $u^{\text{new}}$, i.e., $u^{\text{new}}W_{v}=softmax(\gamma\mathbf{Q}\mathbf{K}^{T})\mathbf{V}$, where we write $\sigma\xi_{Q}=\mathbf{Q}$, $\xi_{K}^{T}\Lambda=\mathbf{K}^{T}$ and $\Lambda^{T}\xi_{K}W_{v}=\mathbf{V}$.  Therefore, the update rule for contextual retrieval from HNs can be equivalent to self-attention through simple conversion.  For self-attention, query pattern is $\mathbf{Q}$ and the context pattern is namely $\mathbf{K}$.

\paragraph{Pattern Retrieval}  From the perspective of memory models, ICL can be reinterpreted as retrieving underlying patterns of input based on context vectors $\lambda$ following the update rule. This interpretation is focused on the association among neurons in some middle layer of the model, where the hidden states at each token position may encode some unique information~\citep{alkhamissi2022review}.  Query and context are thus assumed to be encoded into separate vectors.  The retrieval process consists of the following stages. \textbf{(1)} Query vector $\sigma$ and context vectors $\lambda$ are mapped to the associative space through linear transformation with $\xi_{Q}$ and $\xi_{K}$ to reveal underlying patterns.   \textbf{(2)} Then a similarity score between $u$ and $z$ is computed to measure their mutual closeness in the associative space. Dot product is considered as the similarity function for self-attention.  \textbf{(3)} An exponential separation function, i.e., $Softmax$ is computed to stress the prominent context patterns that have higher similarity scores.  \textbf{(4)} After separation, $u^{\text{new}}$ is computed as a weighted sum of context patterns. There can be repetition in context patterns (which means some $z_{i}=z_{j}$). So more frequent context patterns might thus have a larger contribution to the weighted summation.  

\paragraph{Definition 1} (Query-Context Separation) \textit{For $z_{i}$, $\delta :=uz_{i}-uz_{j},\,\text{where }z_{j} \neq \{z_{i}| i\in [1,M]\}$ and $i,j\in [1,M]$.}

We then establish the distinction in similarity scores between context patterns and query patterns as a metric for evaluating the degree of separation between two context patterns with respect to the query pattern. The larger $\delta$ is between $z_{i}$ and some other patterns $z_{j}$, the easier it will be for $z_{i}$ to be matched by the query pattern $u$.  Moreover, we define the pattern retrieval error as $\|f(u)-u^{\star}\|$, where $f$ is the update rule for the query pattern and $u^{\star}$ is the corresponding underlying ground-truth pattern of $\mathrm{y}$ in the same associative space to $u$. {It is assumed that both the query vectors and context vectors follow some distribution within the pre-trained model, allowing the model to effectively capture and represent their patterns.} {Different from the defined error of HNs in \cite{ramsauer2020hopfield}}, we consider a general case where $u^{T}$ is not necessarily in $\{z_{i}|i \in [1,M]\}.$

\subsection{Prompting Performance as Retrieval Error}
\paragraph{Theorem 1}(Retrieval Error)\label{theo1} \textit{For some $z_{i}$ that has $t$ instances, i.e., $t=\sum_{j=1}^{M}\mathbbm{1}\{z_{j}=z_{i}\}$. The ground-truth pattern $u^{\star}=(z_{i}+\Delta z)^{T}$. We define $c:=\text{exp}(-\gamma (uz_{i}-\text{max}_{z_{i} \neq z_{j}}uz_{j}))=\text{exp}(-\gamma \delta_{min})$, and $z_{max}=\text{max}(z_{1},...,z_{M})$.  The retrieval error $\epsilon:=\|f(u)-u^{\star}\|$ is then bounded by} $\left [0,\,\|\Delta z\|+\beta \|z_{max}\|\right ]$, where $\beta=\left(1-{\left(1+\frac{c(M-t)}{t}\right)}^{-1}+ c(M-t)\right)$ and $\beta  \propto c\frac{M}{t}$

 The proof is displayed in Appendix~\ref{app:proof}. We can see the upper bound consists of two parts, i.e., $\|\Delta_{z}\|$ and $\beta \|z_{max}\|$. We name $\|\Delta_{z}\|$ as \textbf{Instance Error} which directly reflects the match between a context pattern $z_{i}$ and the target pattern $u^{\star}$.  On the other hand, $\beta \|z_{max}\|$ is named as \textbf{Contextual Error} that mainly indicates the separation of $z_{i}$ from other context patterns (remind that $\beta \propto c\frac{M}{t}=\text{exp}(-\gamma \delta_{min})\frac{M}{t}$), i.e., how easy for the model to rely on $z_{i}$ more for the retrieval.  Additionally, when $t=1$ and $||\Delta z||=0$, we are directly retrieving the pattern from context patterns stored in the HN. We next discuss two primary questions on ICL, utilizing our retrieval framework as a foundation, and offer some theoretical predictions.

\paragraph{How does the relation among context patterns influence retrieval error?}  {We first assume the instance error is already acceptably small, otherwise the decrease of the contextual error in the upper bound can be trivial to the total error.} Then from Theorem 1, given the $\|z_{max}\|$, the contextual error is proportionate to $c\frac{M}{t}$. Recall that $c=\text{exp}(-\gamma (uz_{i}-\text{max}_{z_{i} \neq z_{j}}uz_{j}))=\text{exp}(-\gamma \delta_{min})$. When $\delta_{min}$ is larger, the context pattern $z_{i}$ is well separated from other distinct context patterns for the query pattern $u$.  $z_{i}$ will be more prominent when conducting softmax in Eq.~\ref{eq:2} and the upper bound of $\epsilon$ will be lower, which indicates the potential of smaller error.  

\paragraph{How does the number of context patterns influence retrieval error?}\label{para:theo_k}
With the increase of $M$, $\frac{M}{t}$ and $c$ may change accordingly depending on newly introduced context patterns. Given the instance error, this fluctuation leads to the different tendencies of the upper bound, which means varied potential of the actual error. When context vectors are randomly sampled from the distribution, larger $M$ is often observed to enable generally better performance of ICL~\cite{min-etal-2022-rethinking,brown2020language,ye2023compositional}.  However, chances are that if one context pattern $z_{i}$ already has minimum instance error, larger $M$ may lead to declined performance due to the introduced contextual error from other context patterns (i.e., increased $\frac{M}{t}$ and $c$). We empirically show this in Sec.~\ref{sec:exp_cmp}. Thus, the influence of $M$ can be uncertain depending on chosen context patterns.

\subsection{Implications on Exemplar Selection}
\label{sec:select}
This section analyzes the default random selection based on our retrieval framework.  We first detail the definition of exemplar selection.

\paragraph{Definition 2} (Exemplar Selection) \textit{For an input query $\mathbf{x}$ and output $\mathbf{y}$ sampled from distribution $p(\mathcal{D}^{te})$ of task $\mathcal{T}$, a set of $K$ exemplars $\mathcal{S}_{\textup{context}}$ is selected from training data $\mathcal{D}^{tr}_{(x,y)}$ to minimize $\ell(\mathbf{y},\tilde{\mathbf{y}})$, where $\tilde{\mathbf{y}}=\textup{argmax}_{\mathbf{y}}P(\mathbf{y}|e_{1},...,e_{K},\mathbf{x}),\,e_{i}=(x_{e_{i}},y_{e_{i}}),\,e_{i}\in\mathcal{S}_{\textup{context}}$.}

 We assume $p(\mathcal{D}^{te})\approx p(\mathcal{D}^{tr}_{(x,y)})\approx p^{\ast}$ that is the population distribution.  We also regard patterns as latent variables that underlie string sequences.

\paragraph{Random Selection.}\label{pa:rand} Random selection is the default  method~\cite{brown2020language} that can be considered as sampling exemplars from $p(\mathcal{D}^{tr}_{(x,y)})$. When $K$ is large enough, we assume $p(\mathcal{S}_{\text{context}}) \approx p(\mathcal{D}^{tr}_{(x,y)})\approx 
 p^{\ast} $. Accordingly, the mode of context patterns, i.e., $\textup{argmax}_{z}\sum_{j=1}^{M}\mathbbm{1}\{z_{j}=z\}$ may approximate the mode (denoted by $\hat{z}$) of the pattern distribution of samples from $p^{\ast}$. Then for the upper bound of retrieval error $\epsilon$ with $z_{i}=\textup{argmax}_{z}\sum_{j=1}^{M}\mathbbm{1}\{z_{j}=z\}$, the instance error can be approximated to the error given by $\hat{z}$, i.e., $\|\Delta z\|\approx \|\hat{z}-(u^{\ast})^{T}\|$.   Because $\hat{z}$ may more or less be relevant to $(u^{\ast})^{T}$ when they follow the same pattern distribution, with sufficiently large $K$, random selection may give a decent retrieval error.  On the other hand, when $K$ is small, random selection may perform poorly and have great variance depending on sampled exemplars. We provide empirical verification of our theoretical prediction in Fig.~\ref{fig:parse-cmp} of Sec.~\ref{sec:exp_k}.  

\paragraph{Metric-based Selection.} For this approach, a metric function is employed to measure the closeness of query $\mathbf{x}$ and $\mathbf{x}_{e}$ that is the query in an exemplar and to choose more similar $e$ {accordingly~\cite{sorensen-etal-2022-information,gonen2022demystifying,liu-etal-2022-makes,rubin-etal-2022-learning}}. The choice of the metric function is heuristic. One of such approaches is to measure semantic {closeness~\cite{liu-etal-2022-makes,rubin-etal-2022-learning}}, which assumes that semantically similar queries imply similar patterns as well for a task $\mathcal{T}$. But this assumption is likely to collapse for some tasks (see experiments in Sec.~\ref{sec:exp_ret}).


\subsubsection{Active Exemplar Selection}
\label{sec:active}
As has been discussed, the problem with random selection is its great variance, which requires sufficiently large number of exemplars to be effective. However, the number of exemplars is constrained by token limit of the model, making random selection fail in some cases.  Thus, we are motivated to directly sample exemplars with lower expected value of instance error to reduce the upper bound of $\epsilon$.  Inspired by active learning~\cite{roy2001toward}, we propose Active Exemplar Selection that is task-agnostic, parameter-free and fast at inference.  First of all, a value function is defined for an exemplar $e_{i}$ from $\mathcal{D}_{tr}$, 
\begin{align}
    \mathbf{v}(e_{i})=\mathbf{E}_{(x,y)\sim p^{\ast}} \mathbf{s}(\mathit{F}(e_{i},{x}),{y})
\end{align}
$ \mathbf{s}$ is a task-specific score function (e.g., F1 score) and $F$ stands for LLM.  The value function is meant to measure the expected value of the instance error of patterns at the data level (i.e., assuming $\mathbf{v}(e_{i}) \propto \mathbf{E}\|\Delta z\|$).  However, it is hard to directly compute $\mathbf{v}(e_{i})$, we then use Monte Carlo Method to estimate it in the training dataset.
\begin{equation}
    \mathbf{v}(e_{i})\approx \frac{1}{{N-1}}\sum_{j=1,\,j\neq i}^{{N-1}}[ \mathbf{s}(\mathit{F}(e_{i},{x}_{j}),{y}_{j})],
\end{equation}
where $\mathbf{N}$ is the size of training data $\mathcal{D}_{tr}$ and $({x}_{j},{y}_{j}) \sim p(\mathcal{D}^{tr})$.  {Finally, $K$ exemplars with the highest $\mathbf{v}(e_{i})$ on $\mathcal{D}_{tr}$ are selected as the context.}  In this way, exemplars with high $\mathbf{v}(e_{i})$ will be directly chosen that tend to contain the mode pattern of the distribution.  Therefore, active selection can be less noisy and require smaller $K$ to reach  $\|\Delta z\|\approx \|\hat{z}-(u^{\ast})^{T}\|$ than random selection {(see Fig.~\ref{fig:parse-cmp})}.  To make the algorithm more tractable when $|\mathcal{D}_{tr}|$ is very large, instead of using all data, a sufficiently large pool of candidates for comparison can be formed by random sub-sampling of $\mathcal{D}_{tr}$. All in all, active exemplar selection is expected to be a more reliable replacement of default random selection.

\section{Experiments}
\label{sec:exp}

\subsection{Experimental Setup}

\paragraph{Datasets.}  We conduct experiments on two types of common tasks in NLP, each of which has two different datasets. For constituency parsing, we use the Penn Treebank corpus~\cite{marcus-etal-1993-building} with the standard splits (2-21) for training containing 39832 sentences, 22 for validation, 23 for test). PTB is an often-used benchmark for constituency parsing in English. The corpus is collected from a variety of sources including stories, news, and scientific abstracts. To evaluate our approach in a specific domain, we employ the Colorado Richly Annotated Full-Text (CRAFT) corpus~\cite{Cohen2017TheCR} that consists of biomedical journal articles and contains 21121 sentences in total.  We randomly split the CRAFT corpus into a training set and a test set following the ratio of 6:4.  For question answering, we use challenging MedMCQA~\cite{pal2022medmcqa} and MedQA~\cite{jin2021disease} that involve multiple choices on professional biomedical knowledge. We expect those two datasets are harder for LLM to answer with zero-shot prompting, making the effects of in-context exemplars more noticeable.  

\paragraph{Implementation.} We employ Code-Davinci-002, known as Codex~\cite{chen2021evaluating} and GPT-3.5-turbo, i.e., engine for ChatGPT.   For evaluating parsing tasks, in all cases, we report sentence-level unlabeled parsing F1 that is computed separately for each sentence and then averaged across the dataset. For question answering, we report average accuracy over all test data. In terms of configuration of prompting, we use 20 exemplars for constituency parsing, while 5 exemplars for medical QA given longer context and queries.  We employ GPT-3.5-turbo for QA given its more updated training data and ability of following instructions. For active exemplar selection, we randomly sample 100 cases from training data for our estimation of expectance. The score function is F1 for parsing and accuracy for QA.

\paragraph{Baselines.}  To evaluate the effects of our proposed active exemplar selection on ICL, we consider four baselines for comparison: (1) the default random exemplar selection~\cite{brown2020language} that randomly samples exemplars for each test case; (2) semantics-based approach~\cite{liu-etal-2022-makes,rubin-etal-2022-learning} where the relatedness of an exemplar to the input query is decided by computing distance between embeddings and the exemplars with the closest embedding will be chosen; (3) language-modeling-based approach~\cite{shin-etal-2021-constrained} that employs probability $p(u|u_{i})$ output by a large language model where $u$ is a test input and $u_{i}$ is an exemplar. At inference time, both the LM-based method requires traversing partial or all training exemplars to choose the one giving the highest probability, which can be very time-consuming. (4)  BM25~\cite{robertson2009probabilistic} that is a bag-of-words retrieval function to calculate and select exemplars more relevant to the input query.

\subsection{Results}

\begin{table}[htp]
\centering
\begin{tabular}{ccccc}
\toprule
 \textbf{Dataset}                 & \multicolumn{2}{c}{\textbf{PTB}}        & \multicolumn{2}{c}{\textbf{CRAFT}}      \\ 
  \textbf{Model}            & \textbf{Codex}                 & \textbf{Chat} & \textbf{Codex}                 & \textbf{Chat} \\ \hline
Random           &  71.23                     & 69.68        &    58.44       & 57.23        \\
Semantic   &   72.11                    & 70.02        &      57.36      & 57.11       \\
LM         &   71.02                    & 69.73       &   57.23          & 56.84        \\ \hdashline 
Active Selection  & {73.92}   & {73.61}        & {62.98} & {61.65}        \\ \bottomrule
\end{tabular}
\caption{F1 score on PTB and CRAFT with different methods of exemplars selection. }
\label{tab:parse}
\end{table}

\begin{table}
\centering
\begin{tabular}{ccc}
\toprule
   \textbf{Method\textbackslash Dataset}           & \textbf{MedQA} & \textbf{MedMCQA} \\ \hline
Zero-shot         & 43.45           & 41.59         \\
Random           & 60.01      & {54.59}      \\
BM25             & 60.43      & 56.29      \\
Semantic   & 60.33      & 56.14     \\ \hdashline
Active Selection & {62.51}      & {59.86}     \\ 
\bottomrule
\end{tabular}
\caption{Accuracy on MedMCQA and MedQA with different methods of exemplars selection. For zero-shot prompting, no exemplars are used.}
\label{tab:qa1}
\end{table}

\begin{figure}[t]
    \centering
     \centering
    \includegraphics[scale=0.5]{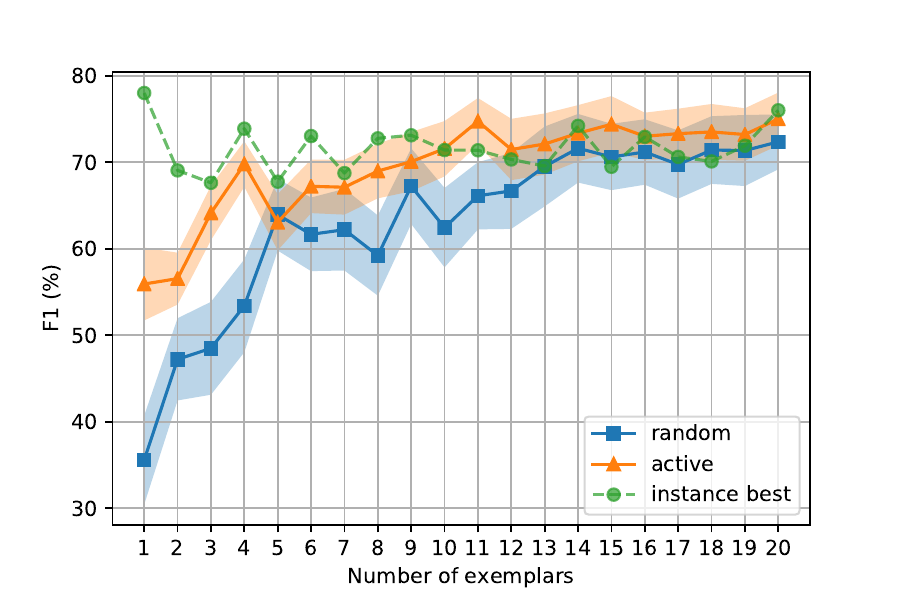}
    \caption{The relation between the performance of ICL with different exemplar selection methods and the number of exemplars (i.e., $K$). Apart from random selection and proposed active selection, we assume the oracle access to ground truth so as to select the best exemplars for each query (denoted as ``instance best'').}
    \label{fig:parse-cmp}
\end{figure}
\paragraph{Downstream Performance.}\label{sec:exp_ret}    The results are shown in Table~\ref{tab:parse} and Table~\ref{tab:qa1}. Our active selection method enables both LLMs to greatly outperform the other baselines. In Table~\ref{tab:parse}, for ChatGPT, the improvement can achieve around 4\%. This demonstrates that choosing useful in-context exemplars help LLM better induce patterns based on its memory.  Additionally, semantic-based selection and LM-based selection give similar performance to the random selection. The assumption of semantic-based selection is that semantically similar data shall contain related task-specific patterns. However, this assumption seems to fail in our evaluation on constituency parsing. In comparison, our proposed active selection requires no such pattern-wise assumption and can directly choose exemplars containing common patterns for a dataset. In Table~\ref{tab:qa1}, the performance on both datasets can be improved by 1\% to 2\% compared with using random exemplars. On MedQA, BM25 and semantic-based selection perform slightly better than random selection. Additionally, using exemplars significantly increases the performance of zero-shot prompting. This demonstrates that LLM may have encoded answers during training and in-context exemplars help increase the accessibility of encoded knowledge through ICL.

\paragraph{Effects of $K$.}\label{sec:exp_k} We conduct experiments with different number of in-context exemplars. The results are shown in Fig.~\ref{fig:parse-cmp}. The performance of ICL increases with more exemplars for random selection and active selection. Recall in our theory in Sec.~\ref{pa:rand}, we consider the drawback of random selection is it needs larger $K$ to reach an optimal status where the mode of exemplars approximates the major patterns of the population. Fig.~\ref{fig:parse-cmp} verifies our reasoning and demonstrates the random selection can achieve a decent performance when $K$ is large enough.  In comparison, our proposed active selection gives much better results even with smaller $K$. This verifies our thoery that active selection can directly capture exemplars that may contain the common patterns. Active selection is thus especially helpful to cases where a large amount of labeled data are available, while the toke limit only allows a few exemplars to be fed to LLM.


\paragraph{Comparison with oracle exemplars.}\label{sec:exp_cmp} For each query instance, we also assume the access to ground truth and rank training data that can give the best performance when used as the only exemplar for the query. We report the average result in Fig.~\ref{fig:parse-cmp}. When $K=1$, such oracle method gives the best result, while the performance drops immediately with additional exemplars and starts to stagnate. This can be due to the increased contextual error.   Including more optimal exemplars turns out to give similar performance to random selection and active selection.   Therefore, different from supervised tuning, for ICL more exemplars do not always guarantee a better performance, which depends on added exemplars.  Additionally, the observation indicates when knowing the optimal exemplar for a query (which is likely to be impossible in practice), we do not need many-shot prompting. However, for cases with no access to such information, simply increasing $K$ may actually be a good strategy for better performance.  

\section{Biological Plausibility of ICL as Memory Retrieval}
\label{sec:bio}
The process of ICL in LLMs exhibits similarities to the memory retrieval process in the human brain, both of which involves the use of prompts or cues related to targeted information to retrieve. This section shows some phenomenon-level biological plausibility of ICL. Similar to LLMs,  human memory retrieval also heavily depends on contextual cues for successful recall~\cite{tulving1973encoding, eich1980cue, godden1975context, smith1979remembering}. 

Human's memories can actually exist in a state of being \textit{available} but \textit{inaccessible}~\cite{tulving1966availability} (compare performance of ``Zero-shot'' with other methods in Table~\ref{tab:qa1}). When some information cannot be recalled with internal cues (i.e., without external hints), such as in free recall tasks, it is considered inaccessible. However, external cues, e.g., category cues related to the target items to recall, can greatly increase the accessibility of memory.  Likewise, LLMs can provide answers to questions that they initially fail in zero-shot prompting scenarios when given related in-context exemplars. The query together with in-context exemplars can also be viewed as partial information cues for memory retrieval, providing incomplete or fragmented versions of the target~\cite{jo2007medial,jo2014memory}.  Additionally, the cue-to-target similarity, also known as encoding specificity, is critical to the likelihood of successful recall for human brain~\cite{tulving1973encoding, nelson1982extralist}. Similarly, LLMs that are trained through language modeling may exhibit such requirements for in-context exemplars~\cite{wang2023robustness}.  

For humans, prompts are typically extralist cues, originating from a different list of stored memories to be retrieved. But extralist cues can still be effective if they are relevant to the target~\cite{tulving1973encoding, nelson1982extralist}. Similarly, in the case of ICL, it is uncommon to encounter context and target output that exactly match the training data. However, by providing relevant exemplars, LLMs may still capture underlying patterns of query with the guide of in-context demonstrations and generalize to unseen cases.

\section{Related Work}
\paragraph{Associative Memory Models.}
Models of associative memory have the capability of recalling its learnt patterns stored in memory for an incomplete input and then recovering it. Hopfield Network (HN)~\citep{hopfield1982neural} was the first introduced type of artificial neural network used for associative memory. It consists of one neural network layer and has binary memories. Memory retrieval is performed through a forward pass following an update rule, which can decrease the energy of the network. An extension of HN is Sparse Distributed Memory (SDM)~\citep{kanerva1988sparse}, which stores memories in an 'Address' matrix and a 'Pattern' matrix as well. Self-attention of Transformer~\citep{vaswani2017attention} is shown to work similarly to SDM~\citep{bricken2021attention}.  Recently, the memory capacity and capabilities of memory models have been raised to a new level. \citet{krotov2016dense} developed Dense Associative Memory for pattern recognition and showed the duality between the associative memory model and a feedforward neural network. \citet{ramsauer2020hopfield} generalized HNs to continus inputs, whose model is known as Modern Hopfield Network (MHN). \citet{ramsauer2020hopfield} demonstrates retrieving patterns from MHN is the same as a feed-forward pass in self-attention~\citep{vaswani2017attention}. On the other hand, the recent work~\citep{hu2023sparse} explores sparse MHN and concurrently, \citet{anonymous2023stanhop} investigate the application of (sparse) Hopfield-based deep learning model to multivariate time series prediction.

\paragraph{In-Context Learning.}
In-Contex Learning (ICL) is the ability of language models to induce answers from given demonstrations without weights updating in supervised tuning. ICL is shown to exist in both small language models (e.g., a vanilla Transformer)~\citep{garg2022can} and large language models~\citep{brown2020language}. Explaining how ICL works for LLM is a fundamental while challenging topic. Currently, there is still no consensus on the mechanism of ICL despite the popularity of LLM in applications. Some of past works~\citep{von2022transformers,dai2022can} propose that ICI is doing implicit gradient descent and theoretically showing that the hidden states of some neurons can be approximately the same as gradients in back-propagation during training. \citet{pan2023context} empirically investigates ICL by disentangling it into task recognition and task learning.  Some other works cast ICL to Bayesian Inference~\citep{xie2022an}.  But those works are mainly tested on a small language model with simple regression tasks, which might not work for LLMs. 

\paragraph{Language Models as Memory Networks.}
Language Models (LMs) have been shown to encode extensive knowledge in their weights~\citep{petroni-etal-2019-language,meng2022locating,roberts-etal-2020-much} through pretraining and can answer factual questions with zero shots or few shots~\citep{li-etal-2022-systematic,singhal2022large}. The output of feed-forward
layer in Transformer is demonstrated to consist of its stored memories~\citep{geva-etal-2021-transformer}. There are works~\citep{wang2022neural,tay2022transformer} that leverage LM for Information Retrieval (IR) by mapping IR tasks to sequence-to-sequence generation tasks and can obtain superior performance. There are also works~\citep{kumar2016ask,daniluk2017frustratingly} that construct memory networks for natural language processing tasks and achieve decent performance as well.

\section{Conclusion}
In this work, we investigate the influence of in-context exemplars on ICL via viewing in-context learning (ICL) of large language models through the lens of models of associative memory. We have shown that in-context learning can be theoretically equivalent to contextual retrieval from a Hopfield Network. We then give theoretical and empirical analysis of formulating exemplars for better downstream performance via ICL and propose more efficient Active Exemplar Selection approach. All in all, our work interprets ICL as contextual retrieval from memory which enables the theoretical analysis of correlation between exemplars and performance of ICL.  By linking recent LLMs to biologically plausible Hopfield Networks, our work may shed new light on understanding LLMs.

\section{Limitations} 
Although we show some convergences between ICL and HNs, our theoretical framework exhibits a high level of abstraction, primarily focusing on a single layer of self-attention and neglecting the influence of order of exemplars. The connection between our theory and  proposals may not be strong enough. We aim to provide more theoretical and empirical evidence in linking ICL and HN as our future work. This work aims to provide explanations rather than precise predictions regarding the impact of context in ICL and may serve as a framework to conceptually analyze in-context learning of LLMs.

\bibliographystyle{plainnat}
\bibliography{ref2}

\begin{thebibliography}{59}
\providecommand{\natexlab}[1]{#1}
\providecommand{\url}[1]{\texttt{#1}}
\expandafter\ifx\csname urlstyle\endcsname\relax
  \providecommand{\doi}[1]{doi: #1}\else
  \providecommand{\doi}{doi: \begingroup \urlstyle{rm}\Url}\fi

\bibitem[AlKhamissi et~al.(2022)AlKhamissi, Li, Celikyilmaz, Diab, and Ghazvininejad]{alkhamissi2022review}
Badr AlKhamissi, Millicent Li, Asli Celikyilmaz, Mona Diab, and Marjan Ghazvininejad.
\newblock A review on language models as knowledge bases.
\newblock \emph{arXiv preprint arXiv:2204.06031}, 2022.

\bibitem[Anonymous(2023)]{anonymous2023stanhop}
Anonymous.
\newblock {ST}anhop: Sparse tandem hopfield model for memory-enhanced time series prediction.
\newblock In \emph{Submitted to The Twelfth International Conference on Learning Representations}, 2023.
\newblock URL \url{https://openreview.net/forum?id=6iwg437CZs}.
\newblock under review.

\bibitem[Barron et~al.(2020)Barron, Auksztulewicz, and Friston]{barron2020prediction}
Helen~C Barron, Ryszard Auksztulewicz, and Karl Friston.
\newblock Prediction and memory: A predictive coding account.
\newblock \emph{Progress in neurobiology}, 192:\penalty0 101821, 2020.

\bibitem[Bonetti et~al.(2021)Bonetti, Brattico, Carlomagno, Donati, Cabral, Haumann, Deco, Vuust, and Kringelbach]{bonetti2021rapid}
Leonardo Bonetti, Elvira Brattico, Francesco Carlomagno, Giovanni Donati, Joana Cabral, Niels~Trusbak Haumann, Gustavo Deco, Peter Vuust, and Morten~L Kringelbach.
\newblock Rapid encoding of musical tones discovered in whole-brain connectivity.
\newblock \emph{NeuroImage}, 245:\penalty0 118735, 2021.

\bibitem[Bricken and Pehlevan(2021)]{bricken2021attention}
Trenton Bricken and Cengiz Pehlevan.
\newblock Attention approximates sparse distributed memory.
\newblock \emph{Advances in Neural Information Processing Systems}, 34:\penalty0 15301--15315, 2021.

\bibitem[Brown et~al.(2020)Brown, Mann, Ryder, Subbiah, Kaplan, Dhariwal, Neelakantan, Shyam, Sastry, Askell, et~al.]{brown2020language}
Tom Brown, Benjamin Mann, Nick Ryder, Melanie Subbiah, Jared~D Kaplan, Prafulla Dhariwal, Arvind Neelakantan, Pranav Shyam, Girish Sastry, Amanda Askell, et~al.
\newblock Language models are few-shot learners.
\newblock \emph{Advances in neural information processing systems}, 33:\penalty0 1877--1901, 2020.

\bibitem[Bubeck et~al.(2023)Bubeck, Chandrasekaran, Eldan, Gehrke, Horvitz, Kamar, Lee, Lee, Li, Lundberg, et~al.]{bubeck2023sparks}
S{\'e}bastien Bubeck, Varun Chandrasekaran, Ronen Eldan, Johannes Gehrke, Eric Horvitz, Ece Kamar, Peter Lee, Yin~Tat Lee, Yuanzhi Li, Scott Lundberg, et~al.
\newblock Sparks of artificial general intelligence: Early experiments with gpt-4.
\newblock \emph{arXiv preprint arXiv:2303.12712}, 2023.

\bibitem[Chen et~al.(2021)Chen, Tworek, Jun, Yuan, Pinto, Kaplan, Edwards, Burda, Joseph, Brockman, et~al.]{chen2021evaluating}
Mark Chen, Jerry Tworek, Heewoo Jun, Qiming Yuan, Henrique Ponde de~Oliveira Pinto, Jared Kaplan, Harri Edwards, Yuri Burda, Nicholas Joseph, Greg Brockman, et~al.
\newblock Evaluating large language models trained on code.
\newblock \emph{arXiv preprint arXiv:2107.03374}, 2021.

\bibitem[Cohen et~al.(2017)Cohen, Verspoor, Fort, Funk, Bada, Palmer, and Hunter]{Cohen2017TheCR}
Kevin~Bretonnel Cohen, Karin~M. Verspoor, Kar{\"e}n Fort, Christopher~S. Funk, Michael Bada, Martha Palmer, and Lawrence~E. Hunter.
\newblock The colorado richly annotated full text (craft) corpus: Multi-model annotation in the biomedical domain.
\newblock 2017.

\bibitem[Dai et~al.(2022)Dai, Sun, Dong, Hao, Sui, and Wei]{dai2022can}
Damai Dai, Yutao Sun, Li~Dong, Yaru Hao, Zhifang Sui, and Furu Wei.
\newblock Why can gpt learn in-context? language models secretly perform gradient descent as meta optimizers.
\newblock \emph{arXiv preprint arXiv:2212.10559}, 2022.

\bibitem[Daniluk et~al.(2017)Daniluk, Rockt{\"a}schel, Welbl, and Riedel]{daniluk2017frustratingly}
Micha{\l} Daniluk, Tim Rockt{\"a}schel, Johannes Welbl, and Sebastian Riedel.
\newblock Frustratingly short attention spans in neural language modeling.
\newblock \emph{arXiv preprint arXiv:1702.04521}, 2017.

\bibitem[Dong et~al.(2022)Dong, Li, Dai, Zheng, Wu, Chang, Sun, Xu, and Sui]{dong2022survey}
Qingxiu Dong, Lei Li, Damai Dai, Ce~Zheng, Zhiyong Wu, Baobao Chang, Xu~Sun, Jingjing Xu, and Zhifang Sui.
\newblock A survey for in-context learning.
\newblock \emph{arXiv preprint arXiv:2301.00234}, 2022.

\bibitem[Eich(1980)]{eich1980cue}
James~Eric Eich.
\newblock The cue-dependent nature of state-dependent retrieval.
\newblock \emph{Memory \& Cognition}, 8:\penalty0 157--173, 1980.

\bibitem[Garg et~al.(2022)Garg, Tsipras, Liang, and Valiant]{garg2022can}
Shivam Garg, Dimitris Tsipras, Percy~S Liang, and Gregory Valiant.
\newblock What can transformers learn in-context? a case study of simple function classes.
\newblock \emph{Advances in Neural Information Processing Systems}, 35:\penalty0 30583--30598, 2022.

\bibitem[Geva et~al.(2021)Geva, Schuster, Berant, and Levy]{geva-etal-2021-transformer}
Mor Geva, Roei Schuster, Jonathan Berant, and Omer Levy.
\newblock Transformer feed-forward layers are key-value memories.
\newblock In \emph{Proceedings of the 2021 Conference on Empirical Methods in Natural Language Processing}. Association for Computational Linguistics, 2021.

\bibitem[Godden and Baddeley(1975)]{godden1975context}
Duncan~R Godden and Alan~D Baddeley.
\newblock Context-dependent memory in two natural environments: On land and underwater.
\newblock \emph{British Journal of psychology}, 66\penalty0 (3):\penalty0 325--331, 1975.

\bibitem[Gonen et~al.(2022)Gonen, Iyer, Blevins, Smith, and Zettlemoyer]{gonen2022demystifying}
Hila Gonen, Srini Iyer, Terra Blevins, Noah~A Smith, and Luke Zettlemoyer.
\newblock Demystifying prompts in language models via perplexity estimation.
\newblock \emph{arXiv preprint arXiv:2212.04037}, 2022.

\bibitem[Hopfield(1982)]{hopfield1982neural}
John~J Hopfield.
\newblock Neural networks and physical systems with emergent collective computational abilities.
\newblock \emph{Proceedings of the national academy of sciences}, 79\penalty0 (8):\penalty0 2554--2558, 1982.

\bibitem[Hu et~al.(2023)Hu, Yang, Wu, Xu, Chen, and Liu]{hu2023sparse}
Jerry Yao-Chieh Hu, Donglin Yang, Dennis Wu, Chenwei Xu, Bo-Yu Chen, and Han Liu.
\newblock On sparse modern hopfield model.
\newblock \emph{arXiv preprint arXiv:2309.12673}, 2023.

\bibitem[Jin et~al.(2021)Jin, Pan, Oufattole, Weng, Fang, and Szolovits]{jin2021disease}
Di~Jin, Eileen Pan, Nassim Oufattole, Wei-Hung Weng, Hanyi Fang, and Peter Szolovits.
\newblock What disease does this patient have? a large-scale open domain question answering dataset from medical exams.
\newblock \emph{Applied Sciences}, 11\penalty0 (14):\penalty0 6421, 2021.

\bibitem[Jo and Choi(2014)]{jo2014memory}
Yong~Sang Jo and June-Seek Choi.
\newblock Memory retrieval in response to partial cues requires nmda receptor-dependent neurotransmission in the medial prefrontal cortex.
\newblock \emph{Neurobiology of learning and memory}, 109:\penalty0 20--26, 2014.

\bibitem[Jo et~al.(2007)Jo, Park, Kim, Park, Kim, Kim, and Choi]{jo2007medial}
Yong~Sang Jo, Eun~Hye Park, Il~Hwan Kim, Soon~Kwon Park, Hyun Kim, Hyun~Taek Kim, and June-Seek Choi.
\newblock The medial prefrontal cortex is involved in spatial memory retrieval under partial-cue conditions.
\newblock \emph{Journal of Neuroscience}, 27\penalty0 (49):\penalty0 13567--13578, 2007.

\bibitem[Kaiser and Bengio(2016)]{kaiser2016can}
{\L}ukasz Kaiser and Samy Bengio.
\newblock Can active memory replace attention?
\newblock \emph{Advances in Neural Information Processing Systems}, 29, 2016.

\bibitem[Kanerva(1988)]{kanerva1988sparse}
Pentti Kanerva.
\newblock \emph{Sparse distributed memory}.
\newblock MIT press, 1988.

\bibitem[Krotov and Hopfield(2016)]{krotov2016dense}
Dmitry Krotov and John~J Hopfield.
\newblock Dense associative memory for pattern recognition.
\newblock \emph{Advances in neural information processing systems}, 29, 2016.

\bibitem[Kumar et~al.(2016)Kumar, Irsoy, Ondruska, Iyyer, Bradbury, Gulrajani, Zhong, Paulus, and Socher]{kumar2016ask}
Ankit Kumar, Ozan Irsoy, Peter Ondruska, Mohit Iyyer, James Bradbury, Ishaan Gulrajani, Victor Zhong, Romain Paulus, and Richard Socher.
\newblock Ask me anything: Dynamic memory networks for natural language processing.
\newblock In \emph{International conference on machine learning}, pages 1378--1387. PMLR, 2016.

\bibitem[Li et~al.(2022)Li, Kuncoro, Hoffmann, de~Masson~d{'}Autume, Blunsom, and Nematzadeh]{li-etal-2022-systematic}
Xiang~Lorraine Li, Adhiguna Kuncoro, Jordan Hoffmann, Cyprien de~Masson~d{'}Autume, Phil Blunsom, and Aida Nematzadeh.
\newblock A systematic investigation of commonsense knowledge in large language models.
\newblock In \emph{Proceedings of the 2022 Conference on Empirical Methods in Natural Language Processing}. Association for Computational Linguistics, 2022.

\bibitem[Liu et~al.(2022)Liu, Shen, Zhang, Dolan, Carin, and Chen]{liu-etal-2022-makes}
Jiachang Liu, Dinghan Shen, Yizhe Zhang, Bill Dolan, Lawrence Carin, and Weizhu Chen.
\newblock What makes good in-context examples for {GPT}-3?
\newblock In \emph{Proceedings of Deep Learning Inside Out (DeeLIO 2022): The 3rd Workshop on Knowledge Extraction and Integration for Deep Learning Architectures}. Association for Computational Linguistics, 2022.

\bibitem[Marcus et~al.(1993)Marcus, Santorini, and Marcinkiewicz]{marcus-etal-1993-building}
Mitchell~P. Marcus, Beatrice Santorini, and Mary~Ann Marcinkiewicz.
\newblock Building a large annotated corpus of {E}nglish: The {P}enn {T}reebank.
\newblock \emph{Computational Linguistics}, 19\penalty0 (2), 1993.

\bibitem[Meng et~al.(2022)Meng, Bau, Andonian, and Belinkov]{meng2022locating}
Kevin Meng, David Bau, Alex Andonian, and Yonatan Belinkov.
\newblock Locating and editing factual associations in {GPT}.
\newblock \emph{Advances in Neural Information Processing Systems}, 36, 2022.

\bibitem[Millidge et~al.(2022)Millidge, Salvatori, Song, Lukasiewicz, and Bogacz]{millidge2022universal}
Beren Millidge, Tommaso Salvatori, Yuhang Song, Thomas Lukasiewicz, and Rafal Bogacz.
\newblock Universal hopfield networks: A general framework for single-shot associative memory models.
\newblock In \emph{International Conference on Machine Learning}, pages 15561--15583. PMLR, 2022.

\bibitem[Min et~al.(2022)Min, Lyu, Holtzman, Artetxe, Lewis, Hajishirzi, and Zettlemoyer]{min-etal-2022-rethinking}
Sewon Min, Xinxi Lyu, Ari Holtzman, Mikel Artetxe, Mike Lewis, Hannaneh Hajishirzi, and Luke Zettlemoyer.
\newblock Rethinking the role of demonstrations: What makes in-context learning work?
\newblock In \emph{Proceedings of the 2022 Conference on Empirical Methods in Natural Language Processing}, pages 11048--11064. Association for Computational Linguistics, December 2022.

\bibitem[Nelson et~al.(1982)Nelson, McEvoy, and Friedrich]{nelson1982extralist}
Douglas~L Nelson, Cathy~L McEvoy, and Martha~A Friedrich.
\newblock Extralist cuing and retrieval inhibition.
\newblock \emph{Journal of Experimental Psychology: Learning, Memory, and Cognition}, 8\penalty0 (2):\penalty0 89, 1982.

\bibitem[Pal et~al.(2022)Pal, Umapathi, and Sankarasubbu]{pal2022medmcqa}
Ankit Pal, Logesh~Kumar Umapathi, and Malaikannan Sankarasubbu.
\newblock Medmcqa: A large-scale multi-subject multi-choice dataset for medical domain question answering.
\newblock In \emph{Conference on Health, Inference, and Learning}, pages 248--260. PMLR, 2022.

\bibitem[Pan(2023)]{pan2023context}
Jane Pan.
\newblock \emph{What In-Context Learning “Learns” In-Context: Disentangling Task Recognition and Task Learning}.
\newblock PhD thesis, Princeton University, 2023.

\bibitem[Petroni et~al.(2019)Petroni, Rockt{\"a}schel, Riedel, Lewis, Bakhtin, Wu, and Miller]{petroni-etal-2019-language}
Fabio Petroni, Tim Rockt{\"a}schel, Sebastian Riedel, Patrick Lewis, Anton Bakhtin, Yuxiang Wu, and Alexander Miller.
\newblock Language models as knowledge bases?
\newblock In \emph{Proceedings of the 2019 Conference on Empirical Methods in Natural Language Processing and the 9th International Joint Conference on Natural Language Processing (EMNLP-IJCNLP)}. Association for Computational Linguistics, 2019.

\bibitem[Ramsauer et~al.(2020)Ramsauer, Sch{\"a}fl, Lehner, Seidl, Widrich, Adler, Gruber, Holzleitner, Pavlovi{\'c}, Sandve, et~al.]{ramsauer2020hopfield}
Hubert Ramsauer, Bernhard Sch{\"a}fl, Johannes Lehner, Philipp Seidl, Michael Widrich, Thomas Adler, Lukas Gruber, Markus Holzleitner, Milena Pavlovi{\'c}, Geir~Kjetil Sandve, et~al.
\newblock Hopfield networks is all you need.
\newblock \emph{arXiv preprint arXiv:2008.02217}, 2020.

\bibitem[Ranganath and Knight(2002)]{ranganath2002prefrontal}
Charan Ranganath and Robert~T Knight.
\newblock Prefrontal cortex and episodic memory: Integrating findings from neuropsychology and functional brain imaging.
\newblock \emph{The cognitive neuroscience of memory: Encoding and retrieval}, 1:\penalty0 83, 2002.

\bibitem[Roberts et~al.(2020)Roberts, Raffel, and Shazeer]{roberts-etal-2020-much}
Adam Roberts, Colin Raffel, and Noam Shazeer.
\newblock How much knowledge can you pack into the parameters of a language model?
\newblock In \emph{Proceedings of the 2020 Conference on Empirical Methods in Natural Language Processing (EMNLP)}. Association for Computational Linguistics, 2020.

\bibitem[Robertson et~al.(2009)Robertson, Zaragoza, et~al.]{robertson2009probabilistic}
Stephen Robertson, Hugo Zaragoza, et~al.
\newblock The probabilistic relevance framework: Bm25 and beyond.
\newblock \emph{Foundations and Trends{\textregistered} in Information Retrieval}, 3\penalty0 (4):\penalty0 333--389, 2009.

\bibitem[Roy and McCallum(2001)]{roy2001toward}
Nicholas Roy and Andrew McCallum.
\newblock Toward optimal active learning through monte carlo estimation of error reduction.
\newblock \emph{ICML, Williamstown}, 2:\penalty0 441--448, 2001.

\bibitem[Rubin et~al.(2022)Rubin, Herzig, and Berant]{rubin-etal-2022-learning}
Ohad Rubin, Jonathan Herzig, and Jonathan Berant.
\newblock Learning to retrieve prompts for in-context learning.
\newblock In \emph{Proceedings of the 2022 Conference of the North American Chapter of the Association for Computational Linguistics: Human Language Technologies}. Association for Computational Linguistics, July 2022.

\bibitem[Scott and David(1978)]{scott1978infinite}
Kirkpatrick Scott and Sherrington David.
\newblock Infinite-ranged models of spin-glasses.
\newblock \emph{Physical Review B}, 17\penalty0 (11):\penalty0 4384--4403, 1978.

\bibitem[Shin et~al.(2021)Shin, Lin, Thomson, Chen, Roy, Platanios, Pauls, Klein, Eisner, and Van~Durme]{shin-etal-2021-constrained}
Richard Shin, Christopher Lin, Sam Thomson, Charles Chen, Subhro Roy, Emmanouil~Antonios Platanios, Adam Pauls, Dan Klein, Jason Eisner, and Benjamin Van~Durme.
\newblock Constrained language models yield few-shot semantic parsers.
\newblock In \emph{Proceedings of the 2021 Conference on Empirical Methods in Natural Language Processing}. Association for Computational Linguistics, 2021.

\bibitem[Singhal et~al.(2022)Singhal, Azizi, Tu, Mahdavi, Wei, Chung, Scales, Tanwani, Cole-Lewis, Pfohl, et~al.]{singhal2022large}
Karan Singhal, Shekoofeh Azizi, Tao Tu, S~Sara Mahdavi, Jason Wei, Hyung~Won Chung, Nathan Scales, Ajay Tanwani, Heather Cole-Lewis, Stephen Pfohl, et~al.
\newblock Large language models encode clinical knowledge.
\newblock \emph{arXiv preprint arXiv:2212.13138}, 2022.

\bibitem[Smith(1979)]{smith1979remembering}
Steven~M Smith.
\newblock Remembering in and out of context.
\newblock \emph{Journal of Experimental Psychology: Human Learning and Memory}, 5\penalty0 (5):\penalty0 460, 1979.

\bibitem[Sorensen et~al.(2022)Sorensen, Robinson, Rytting, Shaw, Rogers, Delorey, Khalil, Fulda, and Wingate]{sorensen-etal-2022-information}
Taylor Sorensen, Joshua Robinson, Christopher Rytting, Alexander Shaw, Kyle Rogers, Alexia Delorey, Mahmoud Khalil, Nancy Fulda, and David Wingate.
\newblock An information-theoretic approach to prompt engineering without ground truth labels.
\newblock In \emph{Proceedings of the 60th Annual Meeting of the Association for Computational Linguistics (Volume 1: Long Papers)}. Association for Computational Linguistics, May 2022.

\bibitem[Sukhbaatar et~al.(2015)Sukhbaatar, Weston, Fergus, et~al.]{sukhbaatar2015end}
Sainbayar Sukhbaatar, Jason Weston, Rob Fergus, et~al.
\newblock End-to-end memory networks.
\newblock \emph{Advances in neural information processing systems}, 28, 2015.

\bibitem[Tay et~al.(2022)Tay, Tran, Dehghani, Ni, Bahri, Mehta, Qin, Hui, Zhao, Gupta, et~al.]{tay2022transformer}
Yi~Tay, Vinh Tran, Mostafa Dehghani, Jianmo Ni, Dara Bahri, Harsh Mehta, Zhen Qin, Kai Hui, Zhe Zhao, Jai Gupta, et~al.
\newblock Transformer memory as a differentiable search index.
\newblock \emph{Advances in Neural Information Processing Systems}, 35:\penalty0 21831--21843, 2022.

\bibitem[Tulving and Pearlstone(1966)]{tulving1966availability}
Endel Tulving and Zena Pearlstone.
\newblock Availability versus accessibility of information in memory for words.
\newblock \emph{Journal of verbal learning and verbal behavior}, 5\penalty0 (4):\penalty0 381--391, 1966.

\bibitem[Tulving and Thomson(1973)]{tulving1973encoding}
Endel Tulving and Donald~M Thomson.
\newblock Encoding specificity and retrieval processes in episodic memory.
\newblock \emph{Psychological review}, 80\penalty0 (5):\penalty0 352, 1973.

\bibitem[Vaswani et~al.(2017)Vaswani, Shazeer, Parmar, Uszkoreit, Jones, Gomez, Kaiser, and Polosukhin]{vaswani2017attention}
Ashish Vaswani, Noam Shazeer, Niki Parmar, Jakob Uszkoreit, Llion Jones, Aidan~N Gomez, {\L}ukasz Kaiser, and Illia Polosukhin.
\newblock Attention is all you need.
\newblock \emph{Advances in neural information processing systems}, 30, 2017.

\bibitem[von Oswald et~al.(2022)von Oswald, Niklasson, Randazzo, Sacramento, Mordvintsev, Zhmoginov, and Vladymyrov]{von2022transformers}
Johannes von Oswald, Eyvind Niklasson, Ettore Randazzo, Jo{\~a}o Sacramento, Alexander Mordvintsev, Andrey Zhmoginov, and Max Vladymyrov.
\newblock Transformers learn in-context by gradient descent.
\newblock \emph{arXiv preprint arXiv:2212.07677}, 2022.

\bibitem[Wang et~al.(2023)Wang, Hu, Hou, Chen, Zheng, Wang, Yang, Huang, Ye, Geng, et~al.]{wang2023robustness}
Jindong Wang, Xixu Hu, Wenxin Hou, Hao Chen, Runkai Zheng, Yidong Wang, Linyi Yang, Haojun Huang, Wei Ye, Xiubo Geng, et~al.
\newblock On the robustness of chatgpt: An adversarial and out-of-distribution perspective.
\newblock \emph{arXiv preprint arXiv:2302.12095}, 2023.

\bibitem[Wang et~al.(2022)Wang, Hou, Wang, Miao, Wu, Chen, Xia, Chi, Zhao, Liu, et~al.]{wang2022neural}
Yujing Wang, Yingyan Hou, Haonan Wang, Ziming Miao, Shibin Wu, Qi~Chen, Yuqing Xia, Chengmin Chi, Guoshuai Zhao, Zheng Liu, et~al.
\newblock A neural corpus indexer for document retrieval.
\newblock \emph{Advances in Neural Information Processing Systems}, 35:\penalty0 25600--25614, 2022.

\bibitem[Wei et~al.(2022)Wei, Wang, Schuurmans, Bosma, Chi, Le, and Zhou]{wei2022chain}
Jason Wei, Xuezhi Wang, Dale Schuurmans, Maarten Bosma, Ed~Chi, Quoc Le, and Denny Zhou.
\newblock Chain of thought prompting elicits reasoning in large language models.
\newblock \emph{arXiv preprint arXiv:2201.11903}, 2022.

\bibitem[Xie et~al.(2022)Xie, Raghunathan, Liang, and Ma]{xie2022an}
Sang~Michael Xie, Aditi Raghunathan, Percy Liang, and Tengyu Ma.
\newblock An explanation of in-context learning as implicit bayesian inference.
\newblock In \emph{International Conference on Learning Representations}, 2022.

\bibitem[Ye et~al.(2023)Ye, Wu, Feng, Yu, and Kong]{ye2023compositional}
Jiacheng Ye, Zhiyong Wu, Jiangtao Feng, Tao Yu, and Lingpeng Kong.
\newblock Compositional exemplars for in-context learning.
\newblock \emph{arXiv preprint arXiv:2302.05698}, 2023.

\bibitem[Yoo et~al.(2022)Yoo, Kim, Kim, Cho, Jo, Lee, Lee, and Kim]{DBLP:conf/emnlp/YooKKCJLLK22}
Kang~Min Yoo, Junyeob Kim, Hyuhng~Joon Kim, Hyunsoo Cho, Hwiyeol Jo, Sang{-}Woo Lee, Sang{-}goo Lee, and Taeuk Kim.
\newblock Ground-truth labels matter: {A} deeper look into input-label demonstrations.
\newblock In \emph{Proceedings of the 2022 Conference on Empirical Methods in Natural Language Processing, {EMNLP} 2022}. Association for Computational Linguistics, 2022.

\end{thebibliography}
\appendix
\onecolumn

\section{Proof of Theorem 1}
\label{app:proof}
\textit{Proof}:
\begin{align*}
    \|f(u)-u^{\star}\|&=\|\Delta z + z_{i}-t[softmax(\gamma\, u,\mathrm{Z})]_{i}z_{i}-\sum^{M}_{j,\,z_{j} \neq z_{i}}[softmax(\gamma u\mathrm{Z})]_{j}z_{j}\|\\
    &=\|\Delta z + \left(1-t\frac{\text{exp}(\gamma uz_{i})}{\sum_{j}^{M}\text{exp}(\gamma uz_{j})}\right) z_{i}-\sum^{M}_{j,\,z_{j} \neq z_{i}}\frac{\text{exp}(\gamma uz_{j})}{\sum_{k}^{M}\text{exp}(\gamma uz_{k})}z_{j}\|\\
    &=\|\Delta z + \left(1-\frac{t}{1+\sum_{j,\,j \neq i}^{M}\text{exp}(\gamma (uz_{j}-uz_{i}))}\right) z_{i}-\sum^{M}_{j,\,z_{j} \neq z_{i}}\frac{\text{exp}(\gamma (uz_{j}-uz_{i}))}{1+\sum_{k,\,k \neq i}^{M}\text{exp}(\gamma (uz_{k}-uz_{i}))}z_{j}\|
\end{align*}

For $z_{i}$, $\delta_{min}=uz_{i}-\text{max}_{z_{i} \neq z_{j}}(uz_{j})$ and recall $t=\sum_{j=1}^{M}\mathbbm{1}\{z_{j}=z_{i}\}$, so we can get, 
\begin{align*}
    1-\frac{t}{1+\sum_{j,\,j \neq i}^{M}\text{exp}(\gamma (uz_{j}-uz_{i}))} \le 1-\frac{t}{t+(M-t)\text{exp}(-\gamma \delta_{min})}=1-\frac{1}{1+\frac{M-t}{t}\text{exp}(-\gamma \delta_{min})}.
\end{align*}

For $z_{j}$ and $z_{j} \neq z_{i}$, 
\begin{align*}
    \frac{\text{exp}(\gamma (uz_{j}-uz_{i}))}{1+\sum_{k,\,k \neq i}^{M}\text{exp}(\gamma (uz_{k}-uz_{i}))}\leq \frac{1}{\text{exp}(\gamma \delta_{min})}=\text{exp}(-\gamma \delta_{min}).
\end{align*}
Then, for the retrieval error, we can have,
\begin{align*}
    \epsilon \leq \|\Delta z\|+\left(1-\frac{1}{1+\frac{M-t}{t}\text{exp}(-\gamma \delta_{min})} \right)\|z_{i}\|+ \text{exp}(-\gamma \delta_{min})\sum_{j,\,z_{j}\neq z_{i}}^{M}\|z_{j}\|.
\end{align*}
Let $c:=\text{exp}(-\gamma \delta_{min})$ and $z_{max}=\text{max}(z_{1},...,z_{M})$. Then,
\begin{align*}
    \epsilon \leq \|\Delta z\|+\left(1-\frac{1}{1+\frac{c(M-t)}{t}}+ c(M-t)\right)\|z_{max}\|.\\
   \text{Furthermore, } -\frac{1}{1+\frac{c(M-t)}{t}}+ c(M-t) \propto c\left(\frac{M}{t}+(M-t)\right),\\
   \text{Therefore, }\epsilon \propto c\frac{M}{t},\,\text{given } \|\Delta z\| \text{ and } \|z_{max}\|.
\end{align*}

Thus, we have proved the upper bound of the retrieval error. For the lower bound, if $u^{\star}$ is retrieved without loss, the error will be naturally zero.

\end{document}